\documentclass[runningheads,envcountsame,a4paper]{llncs}

\usepackage{graphicx}
\usepackage{comment}
\usepackage{caption}
\usepackage{amsmath}
\usepackage{booktabs}
\usepackage{diagbox}
\usepackage{xcolor}
\usepackage{colortbl}
\usepackage{multirow}
\usepackage{CJKutf8}
\usepackage{subfigure}

\usepackage{appendix}
\usepackage{bbding}
\usepackage{pifont}
\usepackage{wasysym}
\usepackage{amssymb}
\usepackage[hyphens]{url}
\usepackage[misc]{ifsym}

\toctitle{Single Model for Influenza Forecasting of Multiple Countries by Multi-task Learning}
\tocauthor{Taichi~Murayama}

\makeatletter
\newcommand{\figcaption}[1]{\def\@captype{figure}\caption{#1}}
\newcommand{\tblcaption}[1]{\def\@captype{table}\caption{#1}}
\makeatother

\begin{document}
\title{Single Model for Influenza Forecasting of Multiple Countries by Multi-task Learning}
\titlerunning{Single Model for Influenza Forecasting of Multiple Countries}
\author{Taichi Murayama(\Letter)\inst{1}\orcidID{0000-0003-1148-711X} \and
Shoko Wakamiya\inst{1}\orcidID{0000-0002-9371-1340} \and
Eiji Aramaki\inst{1}\orcidID{0000-0003-0201-3609}}

\authorrunning{T. Murayama et al.}

\institute{Nara Institute of Science and Technology (NAIST), Japan
\email{\{taichi.murayama.mk1,wakamiya,aramaki\}@is.naist.jp}}

\maketitle              
\begin{abstract}
The accurate forecasting of infectious epidemic diseases such as influenza is a crucial task undertaken by medical institutions.
Although numerous flu forecasting methods and models based mainly on historical flu activity data and online user-generated contents have been proposed in previous studies, no flu forecasting model targeting multiple countries using two types of data exists at present.
Our paper leverages multi-task learning to tackle the challenge of building one flu forecasting model targeting multiple countries; each country as each task.
Also, to develop the flu prediction model with higher performance, we solved two issues; finding suitable search queries, which are part of the user-generated contents, and how to leverage search queries efficiently in the model creation.
For the first issue, we propose the transfer approaches from English to other languages.
For the second issue, we propose a novel flu forecasting model that takes advantage of search queries using an attention mechanism and extend the model to a multi-task model for multiple countries' flu forecasts.
Experiments on forecasting flu epidemics in five countries demonstrate that our model significantly improved the performance by leveraging the search queries and multi-task learning compared to the baselines.


\end{abstract}
\section{Introduction}
The control of infectious diseases is an important task for public health authorities as well as all industry stakeholders worldwide.
Various infectious diseases in addition to COVID-19, which has recently attracted global attention, have had a significant impact on global health and the economy. 
The forecasting of infectious disease epidemics is necessary to execute appropriate measures for their control.
In particular, influenza epidemics, a representative class of severe infectious diseases, leads to 290,000 to 650,000 deaths annually~\cite{who}.
Such instances have motivated public health authorities to forecast the consequences of influenza in different countries.

Many studies relating to flu forecasting models have been conducted for a long time.
In recent years, besides models by leveraging historical flu activity, several models have been proposed to forecast the flu volume by exploiting online user-generated contents (UGCs) such as search query data and social media posts to capture human movements as social sensors~\cite{google_flu,twitter_flu,search1}.
The majority of existing flu forecasting models by leveraging UGCs and historical flu activity data focus on one country or each area in one country.
However, we assume that it is feasible to create a single flu forecasting model targeting multiple countries because the flu time series in each country exhibit strong seasonality and therefore, hold strong similarity.
For example, Pearson correlations of the flu time series in the five countries (US, JP, UK, AU and FR) with different cultures, locations, and languages have a moderate correlation with one another (almost all correlations are over 0.6, refer to Appendix A.)
Moreover, in terms of search queries, which are a representative resource, it has been reported that the user search behaviors for health themes in different countries are similar~\cite{similar_behavior1,similar_behavior2,similar_behavior3}; for example, similar search queries are used when looking for a specific disease.
Thus, it is possible that a single model can achieve sufficient flu forecasting for different countries.
Also, the training of a single model using various flu-related data can capture the nature of flu epidemics in each country by escaping from overfitting, which is caused by a lesser degree of historical data in one country for training~\cite{overfit1,overfit2}.

Our study challenges flu forecasting for various countries with one model as a multi-task problem, which enables two or more tasks to be learned jointly and shares information between the respective tasks.
In other words, we treat each country as each task within the framework of multi-task learning.
Besides, for the development of a flu prediction model with higher performance, we solve two issues; how to find suitable search queries and how to leverage the search queries in the model construction.
We address these issues in the following parts of the paper.

The first issue is how to select queries and keywords in search engines as a resource for flu forecasting.
Many methods using UGCs for forecasting the flu volume have been developed since the emergence of Google Flu~\cite{google_flu}, which demonstrated that the number of search queries capturing human behaviors was a good resource for forecasting.
Certain studies~\cite{correlate1,correlate2,transfer_flu,argo} have depended on ``Google Correlate,'' which returns English search queries that are the most highly correlated to an input time series, for the selection of suitable search queries.
However, this approach cannot be used in many areas (non-English-speaking areas) and it has already been unavailable since December 2019.
Therefore, we discuss a method for selecting search queries in languages other than English to create a flu forecasting model for multiple countries.
In particular, we examine two transfer methods of search queries from English to other languages (Japanese and French): the translation-based method and the combination method of word alignment and time-series correlation (Section 3).

The second issue is how to effectively incorporate search queries into a flu forecasting model.
Two types of data have been applied extensively: historical flu activity data involving the previous year's data (known as ``historical ILI data'')~\cite{historical1,historical2,historical3} and online UGC data~\cite{transfer_flu,twitter_flu,ugc1}, which mainly consist of search query data.
A representative example of simultaneous inputs is the ARGO model~\cite{argo}, which is based on linear regression using the input data of the Google search time series and the historical ILI data. 
The ARGO has exhibited superior results for flu forecasting in the US~\cite{argo_ex}.
However, it has recently been reported that the effect of the search query data in a forecast model is small, and historical ILI data is sufficient as input~\cite{not_effect}.
According to these reports, there remains room for considering how to effectively integrate the search query data, whereas these data have improved the forecasting performance in certain cases.
That is, the simple methods of handling these two resources are insufficient for improving forecasting models.
Furthermore, the overall mutual effect between the historical ILI data and search query data is difficult to capture effectively using existing models, which makes it difficult to extract this effect and apply it to tasks.
To tackle this issue, we propose a model that combines inputs by considering the characteristics of input data.
This approach is based on two aspects: the flu time series exhibits strong seasonality and search query data are useful features for forecasting non-seasonal parts.
Specifically, the search query data are used to forecast the deseasonalized component of flu data by leveraging the attention mechanism~\cite{attention}, which is useful for considering the feature importance (Section 4.2).
Subsequently, we use the model addressing the task as a base and extend it to the flu forecasting model for multiple countries (Section 4.3).

Similar to ours, Zou et al.~\cite{ugc1} proposed a multi-task model based on linear and Gaussian regression to forecast the flu volume in the following two problem settings: several states in the US, and two countries, namely the US and England.
Our multi-task model further develops the above in two aspects: we tackle flu forecasting in five countries, each of which differs in terms of the area or language, and we apply not a simple model such as a statistical model, but our novel neural network-based model for multi-task learning to achieve higher accuracy and long-term forecasting.
Other related studies are discussed in detail in Appendix B.

In summary, we aim to construct a flu forecasting model targeting multiple countries by leveraging multi-task learning while solving two issues as below.
First, to find suitable search queries, we examine the transfer methods of the search queries from English to other languages.
Second, we effectively incorporate the search query data into the model, and propose a novel forecasting model that considers the characteristics of the input data, historical ILI data, and search query data.
The experiments demonstrate that the proposed models and methods achieve the best accuracy among comparative models for forecasting flu epidemics in five countries.

\section{Datasets}
\subsubsection*{ILI rates from health agencies}
We obtained weekly ILI rates, representing the number of ILI cases per 100,000 people in a population, as a measure of ILI activity for the US, Japan, Australia, England, and France from their established syndromic surveillance systems, namely the Centers for Disease Control and Prevention\footnote{https://www.cdc.gov/}, the National Institute of Infectious Diseases\footnote{https://www.niid.go.jp/niid/ja/}, Australian Sentinel Practices Research Network\footnote{https://aspren.dmac.adelaide.edu.au/}, Public Health England\footnote{https://www.gov.uk/government/organisations/public-health-england}, and GPs Sentinelles Network\footnote{https://www.sentiweb.fr/}, respectively.
The England data span from 2013/41st week to 2020/29th week, whereas the others span from 2013/26th week to 2020/29th week.
We denote these countries using the corresponding country codes, namely US, JP, AU, FR, and UK.

\subsubsection*{Search query data}
Time series of weekly search query frequencies were retrieved through Google Trends\footnote{https://trends.google.com} as the UGC data.
The frequency represents the weekly search activity of the queries within a specific region.
The two methods for selecting search queries are described in Section~\ref{method_find}.
The time series of the Google Trends data in the training period were normalized to have a minimum value of zero and maximum value of one (min-max normalization).
The data span was the same as that of the ILI rate data.

\section{Methods for finding search queries}\label{method_find}
We proposed two transfer methods, namely the translation-based and word-alignment and temporal correlation based (WT-based) methods, to explore multilingual search queries using a list of English search queries, which were created in previous research~\cite{ugc1} and placed in a URL\footnote{https://github.com/binzou-ucl/google-flu-mtl}.
As input for the proposed model, we selected the top $L$ English search queries for the US, AU, and UK based on the list, and selected each search query in JP and FR corresponding to the English search query based on these one-to-one query mapping methods.
The usefulness of mapping from English to other languages is described in \cite{similar_behavior2,similar_behavior1,transfer_flu}.
These studies pointed out that the volume movement in the search queries is similar among countries with certain health conditions.

\subsubsection*{Translation-based method: }
This is the simplest trasfer method for the conversion of English into other languages.
To select other languages' queries, we translated English search queries into those of the target language.
We used Google Translate\footnote{https://translate.google.com} for the translation-based method.
For Japanese morphemes, which are not separated by spaces, we divided each morpheme and inserted spaces between them.

\subsubsection*{WT-based method: }
It is possible that the translation-based approach, which simply maps the queries to the target language, will not capture suitable queries.
For example, in Japanese, the abbreviation of influenza, ``flu,'' is translated into ``I-N-FU-LU-E-N-ZA'' and is not translated into the Japanese abbreviation of influenza, ``I-N-FU-LU.''
Moreover, it is difficult to select the suitable orthographical variant, the three categories of which used by the three Japanese writing scripts are applied (kanji-script, hiragana-script, and katakana-script).
We solved these problems using the combination WT-based method, which considers the semantic similarity to the English search queries and temporal similarity to the historical ILI data.

Word alignment is one method that is used for creating cross-lingual word embeddings to compute word similarities in different languages, and is trained using sources of monolingual text with a smaller cross-lingual corpus of aligned text~\cite{word_align}.
This approach can solve the above problems.
For the word alignment, we used the method to learn cross-lingual word embeddings proposed by Zhou et al.~\cite{word_using}.
We needed to prepare word embeddings based on the monolingual text for English and the target languages (Japanese and French).
For this purpose, we obtained the word embedding dataset~\cite{fasttext} learned by \textsf{fasttext} from Wikipedia corpora~\cite{fasttext_}.
Thereafter, we applied these word embeddings to the word alignment method.
To search for words with similar meanings, we used cosine similarity to map each word in the search query, except for prepositions and articles, to the $k$ most similar words in other languages using the common word embedding space created by the word alignment.
The similarity score was represented by $\Theta_{w}$.

Temporal correlation is a method for finding a better search query based on the similarity of the time series of the search queries to the time series of the historical ILI data for the forecast.
It was calculated by the Pearson correlation between the time series of the search query, for which candidates were provided by the word alignment, and the time series of the historical ILI in each country.
The score was represented by $\Theta_{t}$.

The WT-based method selects the search query with the best score in the equation $\Theta_{w} + \Theta_{t}$ corresponding to an English search query.
This is inspired by~\cite{transfer_flu}, which used a similar method of selecting search queries for creating a transfer model of flu forecasts.
Our research differs from the previous research in terms of the motivation whereby we discuss how to find better search queries for flu forecasts.

\section{Building a flu forecasting model for multiple countries}

\subsection{Problem formulation}
Our aim is to forecast the future ILI rates in various countries.
We formulate this problem as a supervised machine learning task.
Let $\textbf{X} = \{x_{t-N+1},...,x_{t-1},x_{t}\} \in \mathbb{R}^{N}$ be a time series of historical ILI data containing $N$ weekly data points.
Let $\textbf{Q} = \{q_{t-N+1},...,q_{t-1},q_{t}\} \in \mathbb{R}^{N \times L}$ be the search query data containing $N$ weekly data points and $L$ queries.
Our model forecasts the true $S$-step-ahead values $\textbf{Y} = \{x_{t+1},...,x_{t+S}\} \in \mathbb{R}^{S}$.
We learn a function $f: \{\textbf{X}, \textbf{Q}\} \rightarrow \textbf{Y}$ that maximizes the prediction accuracy in each country.

\subsection{Model structure}
\label{single_task_model}
\begin{figure*}[t]
        \centering
        \includegraphics[width=14cm]{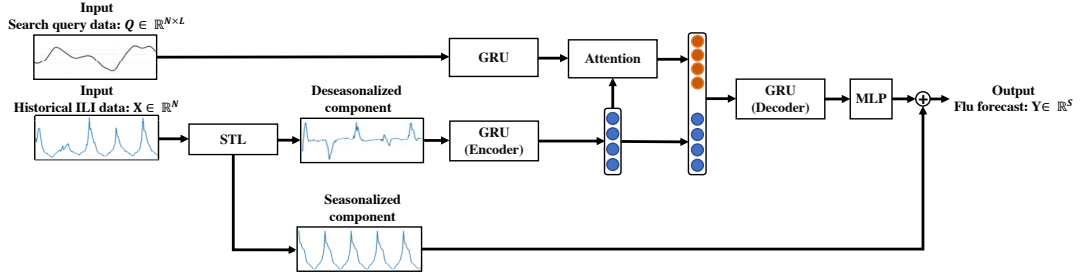}
        \caption{Architecture of proposed model. 
        Historical ILI data are divided into seasonalized and deseasonalized components. 
        We apply the deseasonalized part to the encoder–-decoder model comprising GRUs with an attention mechanism considering search queries.}
        \label{model_simple}
\end{figure*}

Our model is motivated by the idea that search query data are useful features for forecasting non-seasonal parts of flu data.
This concept originates from a previous study~\cite{two_stage}, which reported that the flu forecasting accuracy is improved by splitting the forecasting part from the historical ILI data and search query data.
The model architecture is presented in Fig.~\ref{model_simple}.
For the data preparation, we divide the historical ILI data into the seasonalized and deseasonalized components.
Under the assumption that the seasonalized component has a constant frequency in the future, we forecast the deseasonalized component in the future.
For the forecasting, we apply the encoder–-decoder model considering the search query data using an attention mechanism~\cite{attention}.

\begin{figure*}[t]
        \includegraphics[width=14cm]{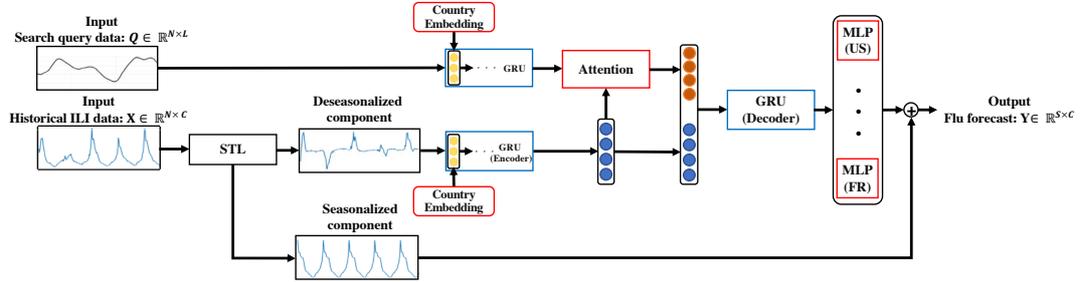}
        \caption{
        Architecture of proposed model expanded to multi-task learning. 
        The red boxes indicate the share of different parameters in the model for each country. 
        The blue boxes indicate the same parameters. Furthermore, country embedding is introduced as the initial latent state of the GRUs.}
        \label{model_multi}
\end{figure*}

\subsubsection*{Flu decomposition: }
We use the seasonal-trend decomposition using LOESS (STL) method~\cite{stl}, which considers the following time-series model with the trend and seasonality: 
$ y_{t} = \tau_{t} + s_{t} + r_{t}, t = 1,2,...,N,$
where $y_{t}$ denotes the historical ILI data at time $t$, $\tau_{t}$ is the trend in the time series, $s_{t}$ is the seasonal signal with period $T$, and $r_{t}$ is the reminder signal.
The seasonal signal describes the repeated patterns in the specified period $T$, which remain constant over time.
The trend describes the continuous increase or decrease.
The detailed decomposition algorithm is outlined in~\cite{stl}.

In our method, the historical ILI data are divided into seasonalized and deseasonalized components.
The deseasonalized component $\textbf{X}^{\tau}$ represents the residual that is obtained by subtracting the seasonalized part $\textbf{X}^{s}$ from the historical ILI data $\textbf{X}$; $\textbf{X}^{\tau} = \textbf{X} - \textbf{X}^{s}$.
Our neural network-based architecture is developed to forecast the future value of a deseasonalized component.
It is assumed that the seasonalized component $\textbf{X}^{s}$ exhibits a constant pattern in the future.
Subsequently, the flu forecast value $\textbf{Y}$ is output by simply adding the value based on the pattern of the seasonalized component to the forecast value of the deseasonalized component using our model.

\subsubsection*{Encoder--decoder model of deseasonalized component: }
Our model employs an encoder–-decoder architecture to forecast more than two weeks ahead.
This architecture is composed of gated recurrent units (GRUs)~\cite{gru}, representing a simple and powerful variant of the RNN, and an attention mechanism, which indicates that the neural network pays close attention to parts of the data when performing tasks.
The GRUs are used to capture the hidden representations of the deseasonalized component of the historical ILI data $\textbf{X}^{\tau}$ and search query data $\textbf{Q}$, and the attention is used to help our model to focus on salient changes in the time series in each query regarding the historical ILI data.
The attention mechanism computes the importance of each query with respect to the forecast and aids in making our model transparent and interpretable.

To capture the hidden representation of the deseasonalized component of the historical ILI data $\textbf{X}^{\tau}$ as the encoder, the GRUs use the input data $\textbf{X}^{\tau}_{t}$ and previous hidden representation $\textbf{H}_{t-1}$, as follows:
\begin{eqnarray}
    \begin{aligned}
        \textbf{r}_{t} = \sigma\left(U_{r}\textbf{X}^{\tau}_{t} + W_{r}\textbf{H}_{t-1}\right), \quad
        \textbf{f}_{t} = tanh\left(U_{h}\textbf{X}^{\tau}_{t} + \textbf{H}_{t-1} \odot W_{h}\textbf{r}_{t}\right), \\
        \textbf{z}_{t} = \sigma\left(U_{z}\textbf{X}^{\tau}_{t} + W_{z}\textbf{H}_{t-1}\right), \quad
        \textbf{H}_{t} = \left(1 - \textbf{z}_{t}\right) \odot \textbf{H}_{t-1} + \textbf{z}_{t} \odot \textbf{f}_{t},
    \end{aligned}
    \label{gru}
\end{eqnarray}
where $\textbf{z}_{t}$ and $\textbf{r}_{t}$ represent the reset and update gates at time $t$, respectively. 
In this case, $U_{z}, U_{r}, U_{h} \in \mathbb{R}^{1 \times M}$, and $W_{z}, W_{r}, W_{h} \in \mathbb{R}^{M \times M}$ are parameters for the respective gates, whereas $M$ is the GRU output dimension.
We combine Equation (\ref{gru}) as follows:
\begin{eqnarray}
    \textbf{H}^{\tau}_{i} = GRU\left(\textbf{X}^{\tau}_{i}\right), \quad i \in \left\{t-N+1, ..., t\right\},
    \label{gru_1}
\end{eqnarray}
where $\textbf{H}^{\tau}_{t} \in \mathbb{R}^{1 \times M}$, which is the last GRU hidden state, is used as the hidden representation.
The search query data also leverage the GRU, as is the case with the historical ILI data.
\begin{eqnarray}
    \textbf{H}^{q}_{i,j} = GRU\left(\textbf{Q}_{i,j}\right), \quad i \in \left\{t-N+1, ..., t\right\}, \quad j \in \left\{1, ..., L\right\},
    \label{gru_2}
\end{eqnarray}
where $\textbf{H}^{q}_{t} \in \mathbb{R}^{L \times M}$, which is the last GRU hidden state, is used as the hidden representation and $L$ is the number of queries.

The combination representation by the attention mechanism is obtained from the hidden representations $\textbf{H}^{\tau}_{t}$ and $\textbf{H}^{q}_{t}$.
In general, an attention mechanism can be defined as mapping a query $q$ and a set of key--value pairs $\{k, v\}$ to an output $o$.
For each position $i$, we compute the attention weighting as the inner product between the query $q_{i}$ and key $k_{i}$ at every position.
For the application of our model, we treat the hidden representation of the deseasonalized component $\textbf{H}^{\tau}_{t}$ as the query, and the hidden representation of the search queries $\textbf{H}^{q}_{t}$ as the key and value.
Position $i$ indicates the location of each search query representation ($i \in \{1,...,L\}$).
The query, key, and value representations are calculated from each representation through linear projection, as follows:
\begin{eqnarray}
    \begin{aligned}
    \textbf{S}_{q} = \textbf{W}^{q}\textbf{H}^{\tau}_{t}, \qquad
    \textbf{S}_{k} = \textbf{W}^{k}\textbf{H}^{q}_{t}, \qquad \textbf{S}_{v} &= \textbf{W}^{v}\textbf{H}^{q}_{t},\\
    \end{aligned}
    \label{attention_eq}
\end{eqnarray}
where $\textbf{S}_{q} \in \mathbb{R}^{1 \times M}$ indicates the query representation, and $\textbf{S}_{k}, \textbf{S}_{v} \in \mathbb{R}^{L \times M}$ indicate the key and value representations, respectively.
Following the linear projection, the dot-product attention computes the importance of each query representation and the attention representation $\textbf{H}^{\tau q}$; $\textbf{H}^{\tau q} = {\rm Softmax}(\textbf{S}_{q}\textbf{S}_{k})\textbf{S}_{v},$
where ${\rm Softmax}(\textbf{S}_{q}\textbf{S}_{k})$ represents the importance of each query and the dimension of $\textbf{H}^{\tau q}$ is $M$.
Thereafter, we apply the feature, concatenating the attention representation $\textbf{H}^{\tau q}$ and hidden representation of the deseasonalized component $\textbf{H}_{t}^{\tau}$, to a multi-layer perceptron (MLP);
$\textbf{H}^{enc} = {\rm MLP}([\textbf{H}_{t}^{\tau} \cdot \textbf{H}^{\tau q}]),$
where the dimension of $\textbf{H}^{enc}$ is $M$.

For the inference of the deseasonalized value of the flu data in the forecast $\{t+1,...,t+S\}$, we apply $\textbf{H}^{enc}$ to the GRU as the decoder and MLP, which constitute two layers.
\begin{align}    
    \begin{cases}
    \textbf{H}^{dec}_{i} &= GRU\left(\textbf{X}^{\tau}_{t}, \textbf{H}^{enc}\right), \quad i = t+1\\
    \textbf{H}^{dec}_{i} &= GRU\left(\textbf{O}^{schedule}_{i-1}\right), \quad i \in \left\{t+2, ..., t+S\right\}
    \label{model_d1} 
    \end{cases}\\
    \hat{\textbf{O}}_{i} = MLP(\textbf{H}^{dec}_{i}), \quad i \in \left\{t+1, ..., t+S\right\},
    \label{model_d2}
\end{align}
where $\hat{\textbf{O}}_{i}$, which is the decoder output, represents the forecast of the deseasonalized value at time $i$.
Moreover, $\textbf{O}^{schedule}_{i-1}$ refers to the value to be applied the scheduled sampling~\cite{schedule}, which is a system of feeding the model with either ground truth values with a probability of $\epsilon$ or forecasts from the model with a probability of $1 - \epsilon$.
This resolves the problem that the discrepancy between the input distributions of the training and testing can lead to poor performance, as the ground truth values are replaced by forecast values generated by the model.

Finally, we can calculate the forecast of the flu volume $\hat{\textbf{Y}}_{i}$ at time $i$ by simply adding the forecast of the deseasonalized value $\hat{\textbf{O}}_{i}$ to the seasonalized value $\textbf{X}^{s}_{i}$;
$\hat{\textbf{Y}}_{i} = \hat{\textbf{O}}_{i} + \textbf{X}^{s}_{i}, i \in \left\{t+1, ..., t+S\right\}$.

\subsubsection*{Training: }
For the model training, we need to determine the true value of the deseasonalized component $\textbf{O}_{i}$.
This is achieved by a simple method, namely subtracting the seasonal part $\textbf{X}^{s}_{i}$ that is assumed to have a constant frequency in the future season from the true flu volume $\textbf{Y}_{i}$.
We use the mean squared error (MSE) loss between the true value of the deseasonalized component $\textbf{O}$ and the forecast value $\hat{\textbf{O}}$.

\subsection{Extension to multi-task model}
\label{multi_task_model}
We extend the proposed model to possess the capability of multi-task learning for flu forecasting in various countries.
Our model aims to improve the expressive ability by means of multi-task learning, which shares part of the learning representations.
The architecture of our multi-task model is presented in Fig.~\ref{model_multi}.
In Fig.~\ref{model_multi}, the components surrounded by blue share all the parameters of the hidden features, whereas those surrounded by red have different parameters set depending on the country.
The GRUs in our model use the same parameters for each task; in particular, parameters of equations (\ref{gru}), (\ref{gru_1}), (\ref{gru_2}), and (\ref{model_d1}) are the same.
The attention and MLP for the final output are set as country-specific; parameters of equations (\ref{attention_eq}) and (\ref{model_d2}) differ for each country.

Furthermore, we propose ``country embedding'' as the initial latent representation of two GRUs regarding the time series of the search queries and deseasonalized component for the multi-task learning of the flu forecasting.
The proposal is based on the possibility of flexible modeling even in the shared representations by changing the initial latent state depending on the forecast target.
The country embedding is calculated as follows:
$\textbf{H}^{country} = {\rm MLP}({\rm Country\_id}),$
where $\textbf{H}^{country} \in \mathbb{R}^{M}$ indicates the initial hidden representation of the GRUs as the input of Equations (\ref{gru_1}) and (\ref{gru_2}), and
``Country\_id'' is the value assigned according to the country (e.g., the ``Country\_id'' of US is 1 and that of JP is 2).
At each step in the training process of the multi-task learning, we randomly select a country, followed by a random training batch $\{\textbf{X}^{country\_id}, \textbf{Q}^{country\_id}, \textbf{Y}^{country\_id}\}$; that is, we set one batch containing only the data of one country at a time.
Our experimental code is public in \url{https://github.com/hkefka385/single-model-for-influenza-forecasting}

\section{Experiments and Results}
\subsection{Experimental settings}
We forecasted the ILI rates in the five countries (US, JP, AU, FR, and UK) using the proposed model.
To validate the forecasting model, the proposed model and other comparative models forecasted the ILI rates from weeks 1 to 5.
We assessed the forecasting performance using three year-long datasets including three flu terms (2017/30th to 2018/29th weeks, 2018/30th to 2019/29th weeks, and 2019/30th to 2020/29th weeks).
We set $52$ weeks (one year) as the validation period before the testing period, and we set more than three years from the initial week of the ILI data to before the validation period as the training period.
We decided to use the WT-based method to identify search queries with all of the models, because the WT-based method is a better approach than translation-based method. (Note that the comparison between the WT-based and translation-based methods is examined in Section \ref{convert_me}.)
We set $52$ weeks as $N$ and $5$ weeks as $S$, which indicated the number of weeks ahead for the forecast.
Furthermore, we set $10$ as $L$, which indicated the number of search queries in the English list as input, and set $100$ as $k$, which indicated the parameter of the WT-based method.
We subsequently selected the learning rate and hidden layer sizes of GRU $M$ as (0.001, 0.01, 0.1, 1.0) and (8, 16, 32, 64), respectively, in the validation period.
During training, all model parameters were updated in a gradient-based manner following the Adam update rule~\cite{adam}.
We set the number of epochs to 300 with early stopping.

We validated the proposed model in the experiments.
\begin{itemize}
    \item \textbf{Proposed w/o sq}: The proposed model was trained using only the historical ILI data of a target country.
    \item \textbf{Proposed\_single}: The proposed model was trained using the data of a target country. The model was the same as that introduced in Section \ref{single_task_model}.
    \item \textbf{Proposed\_multi2}: The proposed model was trained using the data of two target countries, namely the US and JP, for multi-task learning, as in the model introduced in Section \ref{multi_task_model}.
    \item \textbf{Proposed\_multi5}: The proposed model was trained using the data of five target countries for multi-task learning.
\end{itemize}


\subsection{Comparative models}
\begin{itemize}
    \item \textbf{GRU: } The GRU model, one of the recurrent-based models, captures the temporal dependencies in the data and preserves the back-propagated error through the time and layers, referring to equation (\ref{gru}).
    It has been used successfully in influenza forecasting~\cite{lstm}.
    We employed an encoder-–decoder architecture based on the GRU for the multi-step-ahead forecast.
    Two variations of the GRU were used in the experiments: ``GRU w/o sq'' had only historical ILI data, and ``GRU'' had historical ILI and search query data.
    \item \textbf{ARGO: } The ARGO model~\cite{argo} is an autoregressive (AR) model with Google search queries as exogenous variables.
    The simple architecture of this model enables one-step-ahead forecast of the flu volume~\cite{argo3,argo_ex,argo2}.
    The model fails to produce a multi-step-ahead forecast because it requires search query data in advance of the week that we wish to forecast.
    The parameters and input data are the same as those of the proposed model.
    \item \textbf{Transformer: }  The Transformer is one of the most successful models in the NLP.
    Thus far, the Transformer-based flu forecasting model has achieved the highest accuracy~\cite{transfer_flu}.
    \item \textbf{Two-stage: } The Two-stage model~\cite{two_stage}, composed of long short-term model and AR model, was developed inspired by a similar idea to ours, in that the usefulness of the input data differs; historical ILI data and search query data are useful for forecasting the seasonality and trend, respectively. 
    For the multi-step-ahead forecast, we extended the two-stage model to the encoder–-decoder architecture.
    \item \textbf{Multi-task Elastic Net (MTEN): } The MTEN~\cite{ugc1} was proposed as a multi-task model for flu forecasting of US regional areas from search query data.
    This model extends the standard elastic net model to a multi-task version.
    We used the same search queries as those of the proposed model as input. 
    The model outputs a one-step-ahead forecast for the same reason as that of the ARGO model.
    \item \textbf{GRU\_multi: } For the simple comparative method for multi-task, we make one unified model on GRUs, which is trained on the aggregated data of five countries.
    The model is the same setting as GRU, which is one of the comparative methods.
\end{itemize}

To compare the forecast performance levels of each model, we use two evaluation metrics: the coefficient of determination $\rm{R^{2}}$ with a higher value indicating better performance, and root mean squared error $\rm{RMSE}$ with a lower value indicating better performance.

\subsection{Results}
The experimental results for US are presented in Table~\ref{influ_result_us}.
(We examine the experimental results of the other countries in Section \ref{other_result}, and the experimental results for JP in detail are presented in Appendix C.)
This result indicates that the proposed model (particularly our multi-task model) outperformed most baseline methods, confirming the benefits of the model architecture and multi-task learning.
\textbf{GRU w/o sq} and \textbf{GRU} were superior baseline models and achieved approximately $0.8$ to $0.9$ for $\rm{R^{2}}$ in the one-week-ahead forecasts for each country by capturing the temporal dependencies with the RNN architecture.
\textbf{Transformer}, a state-of-the-art flu forecasting method, and \textbf{Two-stage} achieved relatively better scores in the near-ahead forecasts (from 1-week to 3-week) than the GRU-based models, but had almost the same scores in the far-ahead forecasts (from 4-week to 5-week).
These results indicate that it is not easy to improve the accuracy of far-ahead forecasts.
In contrast, the statistical model \textbf{ARGO} achieved relatively lower accuracy than the deep learning models.
We assume that the deep learning-based models were more suitable for flu forecasting in terms of obtaining far-ahead forecast architecture with ease and exhibiting relatively higher accuracy than the statistical-based models, although the calculation cost was high.
Likewise, \textbf{MTEN} based on the statistical model and multi-task learning had the same characteristics.
It tended to exhibit lower accuracy than the other models because its input was only search query data. 
\textbf{GRU\_multi}, a comparavie method for the validation of multi-task, had lower accuracy.
It shows the difficulty of forecasting with a single model without devising model architecture and learning.

Compared to these models, the proposed models (\textbf{Proposed\_single}, \textbf{Proposed\_multi2}, and \textbf{Proposed\_multi5}) achieved the best scores with respect to the terms, metrics, and any-ahead forecasts.
These results reveal that the architecture in the proposed model is useful for flu forecasting.
\textbf{Proposed\_single} achieved the best score among the models without multi-task learning in almost all terms, in which it exhibited the best score in the near-ahead forecast, whereas it had a lower score in the far-ahead forecast than the GRU-based models.

The high degree of the score improvement in \textbf{Proposed\_multi2} and \textbf{Proposed\_multi5} compared to Proposed\_single demonstrated the usefulness of the multi-task learning.
In the near-ahead forecast, the multi-task learning effects were sometimes not observed, whereas the scores of these models in the far-ahead forecast were significantly improved.
For example, in the term 2017 to 2018 in US, the five-week-ahead forecast by Proposed\_multi5 achieved an improvement of 0.136 points in the $\rm{RMSE}$ and 0.059 points in the $\rm{R^{2}}$ compared to Proposed\_single.
Using data from different countries for simultaneous training, the model obtained the latent features of the time series of the ILI rates, thereby improving the forecasting performance.
The difference in the accuracy of the model trained using the data of two countries (Proposed\_multi2) and that trained using the data of five countries (Proposed\_multi5) was not large.

\begin{table*}[!tb]
  \centering
    \caption{Model forecasting performances for US.}
    \small
    \scalebox{0.70}{
\begin{tabular}{c|l|ccc|rrrrrrrrrr} \toprule
     &&& \multicolumn{2}{c|}{Input} &  \multicolumn{2}{c}{1-week} & \multicolumn{2}{c}{2-week} & \multicolumn{2}{c}{3-week} & \multicolumn{2}{c}{4-week} & \multicolumn{2}{c}{5-week}\\ \cmidrule(lr){4-5} \cmidrule(lr){6-7} \cmidrule(lr){8-9} \cmidrule(lr){10-11} \cmidrule(lr){12-13} \cmidrule(lr){14-15}
     Term & \multicolumn{1}{c|}{Model} & Multi & Historical & Query & $\rm{RMSE}$ & $\rm{R^{2}}$ & $\rm{RMSE}$ & $\rm{R^{2}}$ & $\rm{RMSE}$ & $\rm{R^{2}}$ & $\rm{RMSE}$ & $\rm{R^{2}}$ & $\rm{RMSE}$ & $\rm{R^{2}}$ \\ \midrule
     & GRU w/o sq & & \checkmark & & 0.797 & 0.841 & 0.925 & 0.787 & 1.033 & 0.734 & 1.103 & 0.697 & 1.150 & 0.671\\ 
     & Transformer & & \checkmark & & 0.509 & 0.917 & 0.673 & 0.860 & 0.903 & 0.811 & 1.005 & 0.744 & 1.221 & 0.641\\ 
     & *Proposed w/o sq & & \checkmark & & 0.392 & 0.961 & 0.599 & 0.905 & 0.819 & 0.832 & 0.984 & 0.758 & 1.109 & 0.695 \\ 
     & GRU & & \checkmark & \checkmark & 0.783 & 0.849 & 0.905 & 0.791 & 1.025 & 0.741 & 1.097 & 0.705 & 1.138 & 0.654 \\
     2017/30th & ARGO & & \checkmark & \checkmark & 0.405 & 0.954 & --- & --- & --- & --- & --- & --- & --- & --- \\
     -- & Two-stage & & \checkmark & \checkmark & 0.450 & 0.938 & 0.667 & 0.879 & 0.849 & 0.825 & 0.977 & 0.752 & 1.405 & 0.527\\
     2018/29th& *Proposed\_single & & \checkmark & \checkmark & 0.323 & 0.973 & 0.558 & 0.922 & 0.770 & 0.849 & 0.947 & 0.776 & 1.078 & 0.711\\
     & MTEN & \checkmark & & \checkmark & 0.450 & 0.934 & --- & --- & --- & --- & --- & --- & --- & --- \\
     & GRU\_multi & \checkmark & \checkmark & \checkmark & 0.284 & 0.956 & 0.665 & 0.863 & 0.826 & 0.695 & 1.078 & 0.595 & 1.233 & 0.651 \\
     & *Proposed\_multi2 & \checkmark & \checkmark & \checkmark & 0.276 & 0.981 & 0.550 & 0.924 & 0.768 & 0.853 & 0.925 & 0.787 & 1.038 & 0.732 \\
     & *Proposed\_multi5 & \checkmark & \checkmark & \checkmark  & \textbf{0.237} & \textbf{0.986} & \textbf{0.498} & \textbf{0.941} & \textbf{0.692} & \textbf{0.837} & \textbf{0.805} & \textbf{0.832} & \textbf{0.942} & \textbf{0.770} \\ \midrule
     & GRU w/o sq & & \checkmark & &  0.305 & 0.945 & 0.400 & 0.906 & 0.451  & 0.880 & 0.496 & 0.855 & 0.546 & 0.811\\
     & Transformer & & \checkmark & & 0.263 & 0.942 & 0.359 & 0.917 & 0.403 & 0.904 & 0.451 & 0.873 & 0.511 & 0.843\\ 
     & *Proposed w/o sq & & \checkmark & & 0.248 & 0.961 & 0.323 & 0.937 & 0.391 & 0.909 & 0.454 & 0.878 & 0.525 & 0.838\\ 
     & GRU & & \checkmark & \checkmark & 0.283 & 0.940 & 0.371 & 0.915 & 0.439 & 0.887 & 0.472 & 0.865 & 0.528 & 0.838\\
     2018/30th & ARGO & & \checkmark & \checkmark & 0.467 & 0.875 & --- & --- & --- & --- & --- & --- & --- & --- \\
     -- & Two-stage & & \checkmark & \checkmark & 0.308 & 0.947 & 0.417 & 0.891 & 0.481 & 0.854 & 0.517 & 0.857 & 0.541 & 0.810\\
     2019/29th& *Proposed\_single & & \checkmark & \checkmark & \textbf{0.201} & \textbf{0.976} & 0.302 & 0.946 & 0.378 & 0.916 & 0.439 & 0.886 & 0.494 & 0.855\\
     & MTEN & \checkmark & & \checkmark & 0.429 & 0.915 & --- & --- & --- & --- & --- & --- & --- & --- \\
     & GRU\_multi & \checkmark & \checkmark & \checkmark & 0.383 & 0.910 & 0.448 & 0.888 & 0.603 & 0.725 & 0.739 & 0.689 & 0.820 & 0.644 \\
     & *Proposed\_multi2 & \checkmark & \checkmark & \checkmark & 0.232 & 0.968 & \textbf{0.268} & \textbf{0.957} & \textbf{0.323} & \textbf{0.938} & \textbf{0.388} & \textbf{0.888} & \textbf{0.454} & \textbf{0.856}\\
     & *Proposed\_multi5 & \checkmark & \checkmark & \checkmark  &  0.255 & 0.963 & 0.296 & 0.941 & 0.369 & 0.915 & 0.404 & 0.877 & 0.499 & 0.843\\ \midrule
     & GRU w/o sq & & \checkmark & & 0.698 & 0.882 & 0.910 & 0.807 & 1.096 & 0.713 & 1.153 & 0.683 & 1.167 & 0.648 \\ 
     & Transformer & & \checkmark & & 0.659 & 0.892 & 0.919 & 0.807 & 1.099 & 0.712 & 1.154 & 0.680 & 1.218 & 0.652\\ 
     & *Proposed w/o sq & & \checkmark & & 0.538 & 0.932 & 0.838 & 0.837 & 1.084 & 0.725 & 1.171 & 0.683 & 1.241 & 0.605\\ 
     & GRU & & \checkmark & \checkmark & 0.705 & 0.870 & 0.925 & 0.791 & 1.108 & 0.702 & 1.165 & 0.681 & 1.176 & 0.620\\
     2019/30th & ARGO & & \checkmark & \checkmark & 0.984 & 0.758 & --- & --- & --- & --- & --- & --- & --- & ---\\
     -- & Two-stage & & \checkmark & \checkmark & 0.602 & 0.918 & 0.924 & 0.817 & 1.081 & 0.726 & 1.231 & 0.609 & 1.224 & 0.590\\
     2020/29th & *Proposed\_single & & \checkmark & \checkmark & 0.469 & 0.949 & 0.694 & 0.881 & 0.809 & 0.846 & 0.863 & 0.824 & 0.918 & 0.799\\
     & MTEN & \checkmark & & \checkmark & 0.992 & 0.760 & --- & --- & --- & --- & --- & --- & --- & --- \\
     & GRU\_multi & \checkmark & \checkmark & \checkmark & 0.724 & 0.889 & 0.980 & 0.740 & 1.185 & 0.635 & 1.304 & 0.589 & 1.305 & 0.590 \\
     & *Proposed\_multi2 & \checkmark & \checkmark & \checkmark  & 0.409 & 0.961 & 0.641 & 0.904 & 0.770 & 0.861 & 0.840 & 0.833 & 0.910 & 0.802\\
     & *Proposed\_multi5 & \checkmark & \checkmark & \checkmark  & \textbf{0.370} & \textbf{0.971} & \textbf{0.605} & \textbf{0.920} & \textbf{0.696} & \textbf{0.878} & \textbf{0.787} & \textbf{0.853} & \textbf{0.831} & \textbf{0.841}\\
     \bottomrule
  \end{tabular}
  }
  \\\textbf{*} indicates the variation in the proposed model. Bold indicates the best score in each metric and each term.
    \label{influ_result_us}
\end{table*}

\begin{table*}[!tb]
    \centering
    \caption{Forecasting performances of each model for ILI rates in JP, UK, AU, and FR from 2017/30th week to 2018/29th week.}
    \small
    \centering
    \begin{tabular}{c|l|rrrrrrrrrr} \toprule
    \multirow{2}{*}{Country} &  \multicolumn{1}{c|}{\multirow{2}{*}{Model}} & \multicolumn{2}{c}{1-week} & \multicolumn{2}{c}{2-week} & \multicolumn{2}{c}{3-week} & \multicolumn{2}{c}{4-week} & \multicolumn{2}{c}{5-week}\\ \cmidrule(lr){3-4} \cmidrule(lr){5-6} \cmidrule(lr){7-8} \cmidrule(lr){9-10} \cmidrule(lr){11-12}
    & & $\rm{RMSE}$ & $\rm{R^{2}}$ & $\rm{RMSE}$ & $\rm{R^{2}}$ & $\rm{RMSE}$ & $\rm{R^{2}}$ & $\rm{RMSE}$ & $\rm{R^{2}}$ & $\rm{RMSE}$ & $\rm{R^{2}}$ \\ \midrule
    & GRU & 3.412 & 0.939 & 4.019 & 0.923 & 5.223 & 0.915 & 5.982 & 0.826 & 6.164 & 0.813\\
    \multirow{2}{*}{JP} & Proposed\_single & 2.517 & 0.964 & 3.218 & 0.944 & 3.688 & 0.934 & 4.898 & 0.884 & 5.822 & 0.836\\
    & GRU\_multi & 3.261 & 0.944 & 3.901 & 0.882 & 5.072 & 0.820 & 6.552 & 0.776 & 6.752 & 0.756\\
    & Proposed\_multi5 & \textbf{2.429} & \textbf{0.970} & \textbf{2.878} & \textbf{0.951} & \textbf{3.411} & \textbf{0.941} & \textbf{4.057} & \textbf{0.920} & \textbf{4.423} & \textbf{0.905}\\ \midrule
    & GRU & 1.900 & 0.910 & 2.639 & 0.809 & 2.738 & 0.794 & 2.783 & \textbf{0.787} & 3.185 & 0.722\\
    \multirow{2}{*}{UK} & Proposed\_single & 1.794 & 0.912 & 2.591 & 0.816 & 2.959 & 0.770 & 2.901 & 0.741 & 3.100 & 0.729\\
    & GRU\_multi & 6.080 & 0.757 & 8.751 & 0.629 & 9.764 & 0.538 & 10.546 & 0.461 & 10.797 & 0.435 \\
    & Proposed\_multi5 & \textbf{1.510} & \textbf{0.935} & \textbf{2.199} & \textbf{0.873} & \textbf{2.675} & \textbf{0.808} & \textbf{2.709} & 0.783 & \textbf{2.992} & \textbf{0.745}\\ \midrule
    & GRU & 1.754 & 0.939 & 2.085 & 0.914 & 2.430 & 0.884 & 2.739 & 0.852 & 3.122 & 0.807\\
    \multirow{2}{*}{AU} & Proposed\_single & 1.764 & 0.938 & 2.131 & 0.922 & 2.480 & 0.883 & 2.683 & 0.859 & 3.058 & 0.816\\
    & GRU\_multi & 2.458 & 0.933 & 3.099 & 0.876 & 4.080 & 0.821 & 4.674 & 0.765 & 5.018 & 0.728 \\
    & Proposed\_multi5 & \textbf{1.650} & \textbf{0.942} & \textbf{1.999} & \textbf{0.928} & \textbf{2.391} & \textbf{0.899} & \textbf{2.592} & \textbf{0.879} & \textbf{2.794} & \textbf{0.849}\\ \midrule
    & GRU & 0.283 & 0.868 & 0.427 & 0.675 & 0.521 & 0.517 & 0.565 & 0.434 & 0.587 & 0.391\\
    \multirow{2}{*}{FR} & Proposed\_single & 0.266 & 0.874 & 0.413 & 0.696 & 0.507 & 0.542 & 0.551 & 0.461 & 0.560 & 0.443\\
    & GRU\_multi & 0.377 & 0.883 & 0.500 & 0.601 & 0.791 & 0.333 & 0.945 & 0.170 & 1.180 & 0.067\\
    & Proposed\_multi5 & \textbf{0.234} & \textbf{0.904} & \textbf{0.375} & \textbf{0.751} & \textbf{0.452} & \textbf{0.611} & \textbf{0.527} & \textbf{0.511} & \textbf{0.552} & \textbf{0.466}\\ \bottomrule
    \end{tabular}
    \label{country_result}
\end{table*}

\begin{table*}[!tb]
    \centering
    \caption{Comparison of forecasting performances of translation-based and WT-based methods using Proposed\_single model in JP and FR from 2017/30th week to 2018/29th week.}
    \small
    \centering
    \begin{tabular}{c|l|rrrrrrrrrr} \toprule
    \multicolumn{1}{c|}{\multirow{2}{*}{Country}} & \multicolumn{1}{c|}{\multirow{2}{*}{Method}} & \multicolumn{2}{c}{1-week} & \multicolumn{2}{c}{2-week} & \multicolumn{2}{c}{3-week} & \multicolumn{2}{c}{4-week} & \multicolumn{2}{c}{5-week}\\ \cmidrule(lr){3-4} \cmidrule(lr){5-6} \cmidrule(lr){7-8} \cmidrule(lr){9-10} \cmidrule(lr){11-12}
    & & $\rm{RMSE}$ & $\rm{R^{2}}$ & $\rm{RMSE}$ & $\rm{R^{2}}$ & $\rm{RMSE}$ & $\rm{R^{2}}$ & $\rm{RMSE}$ & $\rm{R^{2}}$ & $\rm{RMSE}$ & $\rm{R^{2}}$ \\ \midrule
    \multirow{2}{*}{JP} & Translation-based & \textbf{2.492} & \textbf{0.964} & 3.307 & 0.939 & 3.770 & 0.929 & \textbf{4.800} & 0.880 & 5.976 & 0.828\\
    & WT-based & 2.517 & \textbf{0.964} & \textbf{3.218} & \textbf{0.944} & \textbf{3.688} & \textbf{0.934} & 4.898 & \textbf{0.884} & \textbf{5.822} & \textbf{0.836}\\ \midrule
    \multirow{2}{*}{FR} & Translation-based & 0.278 & 0.856 & 0.432 & 0.661 & 0.531 & 0.491 & 0.592 & 0.407 & 0.594 & 0.405 \\
    & WT-based  & \textbf{0.266} & \textbf{0.874} & \textbf{0.413} & \textbf{0.696} & \textbf{0.507} & \textbf{0.542} & \textbf{0.551} & \textbf{0.461} & \textbf{0.560} & \textbf{0.443}\\
    \bottomrule
    \end{tabular}
    \label{words}
\end{table*}

\section{Discussions}
\subsection{Multi-model performance for other countries}
\label{other_result}
Table~\ref{country_result} displays the forecast performances of four models (GRU, Proposed\_single, GRU\_multi, and Proposed\_multi5) for the ILI rates in JP, UK, AU, and FR from 2017/30th week to 2018/29th week.
These results suggest that the multi-task learning model Proposed\_multi5 achieved the best score in almost all ahead forecasts, as well as the flu forecasting result in US.
Multi-task learning is not limited to a numbefr of countries, but can be applied to various countries with different languages and environments, and the experimental results revealed that the multi-task method improved the forecasting performance.

\subsection{Comparison of models without and with search queries}
Recent research relating to flu forecasts~\cite{not_effect} claimed that the effect of search queries is small.
To tackle this problem, our research presents a model with an attention mechanism that effectively considers search queries.
To examine the search queries' effectiveness, we validated the degree of improvement of the two variation models, namely the GRU-based (GRU w/o sq and GRU) and proposed (Proposed w/o sq and Proposed\_single) models, without and with search queries.

The experimental results for the flu forecast in US (Table~\ref{influ_result_us}) indicate that
the change from GRU w/o sq to GRU resulted in an average improvement of $0.007$ points in the $\rm{RMSE}$, and of $0.001$ points in the $\rm{R^{2}}$.
However, the change from Proposed w/o sq to  Propose\_single resulted in an average improvement of $0.091$ points in the $\rm{RMSE}$, and of $0.017$ points in the $\rm{R^{2}}$.
This suggests that the search query data resulted in the GRU-based models, which simply used the search query data as input, exhibits low improvement scores by adding them.
However, the proposed model, with a well-crafted architecture for the search query data input, achieved a significantly improved score.
These results confirm that it is difficult to treat search queries as input for flu forecasting, and it is necessary to contribute to the score improvement by considering the model devices, such as the introduction of an attention mechanism.

\subsection{Analysis of the methods to find search queries}\label{convert_me}
We compared the translation-based and WT-based methods for the selection of search queries.
For comparison, we experimented with the flu forecast from 2017/30th week to 2018/29th week in JP and FR using the Proposed\_single model with the translation-based and WT-based methods.

The results in Table~\ref{words} demonstrate that the WT-based method achieved better scores than the translation-based method in all experimental metrics in FR and most experimental metrics in JP.
However, the degree of improvement in the accuracy was not large. 
For example, for $\rm{R^{2}}$, the two-week-ahead forecast for JP exhibited only a $0.005$ point improvement, and that for FR exhibited only a $0.035$ point improvement.
Our model based on a neural network can consider a small number of search queries as input for efficient calculation, compared to the multi-task model~\cite{ugc1} based on a statistical method that can consider many search queries.
We assume that the architecture of our model, which does not involve a large number of search queries as input, is insignificantly affected by the selection of search queries.
Although the results demonstrated that the WT-based method was superior as the selection method for our flu forecasting model, substantial room for consideration remains, such as which method is better for models dealing with a large number of search queries.

\section{Conclusions}
In this study, we attempted to construct a flu forecasting model targeting multiple countries by leveraging multi-task framework.
Also, we addressed two tasks: finding suitable search queries in languages other than English and leveraging the search query data as input for the forecasting model.
We revealed that the WT-based method is a better approach for the exploration of search queries.
Moreover, we proposed a novel forecasting model considering the characteristics of the input data, historical ILI data and search query data, and demonstrated the usefulness of the model architecture.
Throughout the flu forecasting experiments in multiple countries, the proposed model achieved the highest performance by acquiring the latent features in the flu time series and by treating the task as multi-task learning.

Our experiments demonstrated the feasibility of constructing a flu forecasting model targeting multiple countries and the usefulness of search query data as input for the proposed model.
However, the method of searching for suitable search queries remains a major challenge, which our research has not yet solved.
Although we used the list of English search queries, a method for identifying appropriate search queries without relying on external resources is required.
Moreover, it is necessary to examine a method to apply the proposed flu forecasting model to new infectious diseases from short period data, such as COVID-19, for dealing with a pandemic.

%
%
%
\bibliographystyle{splncs04}
\bibliography{main}

\newpage
\appendix
\section{Correlation between time series of ILI rates in each country}

\begin{figure}[b]
\begin{tabular}{cc}
  \begin{minipage}{.42\textwidth}
  \captionsetup{width=.95\linewidth}
  \centering
  \tblcaption{Pearson correlation between time series of ILI rates in each country (the only time series in AU shifts forward in 22 weeks to match the peak point of the US and AU).}
  \begin{tabular}{|c|c|c|c|c|c|} \toprule
  & US & JP & UK & AU & FR\\ \midrule
  US & \cellcolor{gray} & 0.793 & 0.614 & 0.840 & 0.745\\
  JP & \cellcolor{gray} & \cellcolor{gray} & 0.592 & 0.751 & 0.527\\
  UK & \cellcolor{gray} & \cellcolor{gray} & \cellcolor{gray} & 0.772 & 0.693\\
  AU & \cellcolor{gray} & \cellcolor{gray} & \cellcolor{gray} & \cellcolor{gray} & 0.681\\
  FR & \cellcolor{gray} & \cellcolor{gray} & \cellcolor{gray} & \cellcolor{gray} & \cellcolor{gray}\\ \bottomrule
  \end{tabular}
  \label{corr}
  \end{minipage}
  \begin{minipage}{.58\textwidth}
  \captionsetup{width=.95\linewidth}
	\centering
    \includegraphics[width=8cm]{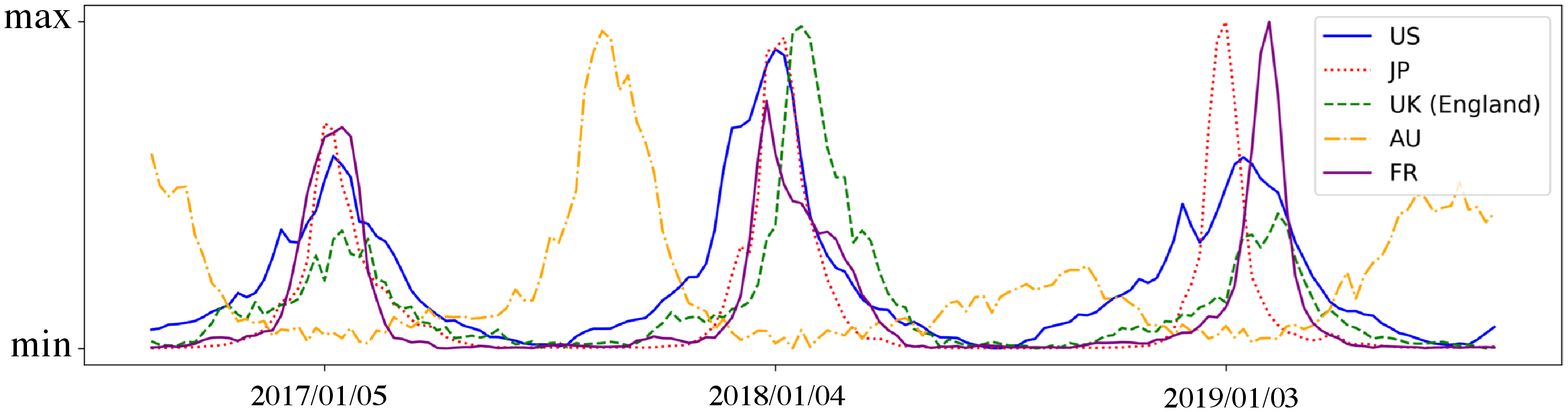}
    \caption{
    Time series of ILI rates in five countries, applied to min-max normalization, from 2016/29th week to 2019/30th week.}
    \label{world}
  \end{minipage}
\end{tabular}
\end{figure}

Fig.~\ref{world} presents the flu time series from 2016/29th week to 2019/30th week in five countries. 
This represents that the flu time series in each country exhibits strong seasonality and therefore, holds strong similarity.
Table~\ref{corr} displays the Pearson correlations of the time series among the respective countries.
These values suggest that the influenza-like illness (ILI) rates in the five countries with different cultures, locations, and languages have a moderate correlation with one another (almost all correlations are over 0.6).
The high correlation gives us a strong motivation to addresses the challenge of flu forecasting for various countries in one model.

\section{Related Work}
Flu forecasting is a type of time-series prediction.
Time-series prediction tasks are mainly divided into univariate and multivariate types~\cite{time_survey}.
In research on time-series prediction, various models suitable for each task have been proposed over time.
Owing to the rapid development of neural networks, many models have been based on these, particularly convolutional neural networks (CNNs)~\cite{wavenet,cnn1,cnn2} and recurrent neural networks (RNNs)~\cite{deepar,rnn_state,rnn_ss2}, which capture the temporal variation.
In recent years, there has been an increase in time-series prediction models using ``attention'' (Transformer) to achieve state-of-the-art performance in multiple natural language processing applications~\cite{attention,transformer_original}.
Attention generally aggregates temporal features using dynamically generated weights, thereby enabling the network to focus on significant time steps in the past directly.
For example,~\cite{attention_apply1,attention_apply2,attention_apply3} are time-series prediction models based on attention.
Although numerous models and methods have been proposed to achieve higher prediction accuracy, it is difficult to apply them to the influenza forecasting problem in a simple manner.
This is owing to the problem setting of flu forecasting; that is, the future flu volume is estimated from two major resources: historical ILI data and UGC data.
This problem belongs to a multivariable problem with one objective variable and various explanatory variables, and most models are not designed for this problem type.
Thus, we need to develop a time-series model suitable for flu and not apply state-of-the-art prediction models.

Numerous models relating to flu forecasting have been proposed to date.
Prior to the emergence of online UGCs, compartmental models such as SIR~\cite{sir} and IDEA~\cite{idea}, as well as statistical models such as autoregressive models~\cite{ar} using historical ILI data, were used extensively for flu forecasting.
With the development of the Internet, some researches~\cite{www_ad1,www_ad2} revealed that online UGCs such as search queries and social media posts are useful resources, same as in the field of influenza prediction~\cite{google_flu}.
Although certain studies have used only one resource of either historical ILI data or online UGCs, the majority of studies proposed supervised methods using online UGCs together with historical ILI data simultaneously as input.
Most of these approaches do not consider the characteristics of the data types, but simply simultaneous inputting, when learning the model.
Several previous models~\cite{argo2,two_stage} based on statistical methods have been developed to exploit the characteristics of the different input data.
However, these studies exhibit certain disadvantages, such as the necessity of long training terms or a small degree of improved accuracy.
Our method based on neural networks to capture the latent features can achieve the best accuracy in flu forecasting by means of an appropriate combination method of two inputs: historical ILI data and search query data.

Moreover, we aim to learn a single flu forecasting model for multiple countries as multi-task learning.
Multi-task learning, which was introduced by~\cite{mtl}, improves the generalization, and achieves superior efficiency and prediction accuracy by using a shared representation from related tasks.
It is used extensively in various areas, such as natural language processing (NLP)~\cite{multi_nlp1,multi_nlp2} and time series~\cite{multi_time1,multi_time2}.
The fundamentals of multi-task learning were presented in detail in~\cite{mtl}.
Zou et al.~\cite{ugc1} tackled a similar problem to ours and proposed a multi-task model based on linear and Gaussian regression to forecast the flu volume in the following two problem settings: several states in the US, and two countries, namely the US and England.
Our multi-task model and task further develop the above in two aspects: we tackle flu forecasting in five countries, each of which differs in terms of the area or language, and we apply not a simple model such as a statistical model, but our novel neural network-based model for multi-task learning to achieve higher accuracy and long-term forecasting.

\begin{table*}[!tb]
  \centering
    \caption{Model forecasting performances for JP.}
    \small
    \scalebox{0.70}{
\begin{tabular}{c|l|ccc|rrrrrrrrrr} \toprule
     &&& \multicolumn{2}{c|}{Input} &  \multicolumn{2}{c}{1-week} & \multicolumn{2}{c}{2-week} & \multicolumn{2}{c}{3-week} & \multicolumn{2}{c}{4-week} & \multicolumn{2}{c}{5-week}\\ \cmidrule(lr){4-5} \cmidrule(lr){6-7} \cmidrule(lr){8-9} \cmidrule(lr){10-11} \cmidrule(lr){12-13} \cmidrule(lr){14-15}
     Term & \multicolumn{1}{c|}{Model} & Multi & Historical & Query & $\rm{RMSE}$ & $\rm{R^{2}}$ & $\rm{RMSE}$ & $\rm{R^{2}}$ & $\rm{RMSE}$ & $\rm{R^{2}}$ & $\rm{RMSE}$ & $\rm{R^{2}}$ & $\rm{RMSE}$ & $\rm{R^{2}}$ \\ \midrule
     & GRU w/o sq& & \checkmark & & 3.998 & 0.922 & 4.876 & 0.885 & 5.313 & 0.863 & 5.836 & 0.835 & 6.275 & 0.810\\ 
     & Transformer & & \checkmark & & 3.217 & 0.949 & 4.202 & 0.936 & 4.864 & 0.895 & 5.715 & 0.849 & 6.353 & 0.819 \\ 
     & *Proposed w/o sq & & \checkmark & & 3.087 & 0.956 & 3.811 & 0.935 & 4.362 & 0.903 & 5.518 & 0.853 & 6.052 & 0.824 \\ 
     2017/30th & GRU & & \checkmark & \checkmark  & 3.412 & 0.939 & 4.019 & 0.923 & 5.223 & 0.915 & 5.982 & 0.826 & 6.164 & 0.813 \\
     --- & ARGO & & \checkmark & \checkmark & 3.317 & 0.941 & --- & --- & --- & --- & --- & --- & --- & --- \\
     2018/29th & Two-stage & & \checkmark & \checkmark & 4.727 & 0.898 & 4.866 & 0.851 & 5.653 & 0.822 & 6.301 & 0.818 & 6.814 & 0.787 \\
     & *Proposed\_single & & \checkmark & \checkmark & 2.517 & 0.964 & 3.218 & 0.944 & 3.688 & 0.934 & 4.898 & 0.884 & 5.822 & 0.836\\
     & MTEN & \checkmark & & \checkmark & 3.697 & 0.922 & --- & --- & --- & --- & --- & --- & --- & --- \\
     & GRU\_multi & \checkmark & \checkmark & \checkmark & 3.261 & 0.944 & 3.901 & 0.882 & 5.072 & 0.820 & 6.552 & 0.776 & 6.752 & 0.756\\
     & *Proposed\_multi2 & \checkmark & \checkmark & \checkmark & 2.665 & 0.965 & 3.163 & \textbf{0.951} & 3.569 & 0.938 & 4.298 & 0.910 & 4.652 & 0.895\\
     & *Proposed\_multi5 & \checkmark & \checkmark & \checkmark  & \textbf{2.429} & \textbf{0.970} & \textbf{2.878} & \textbf{0.951} & \textbf{3.411} & \textbf{0.941} & \textbf{4.057} & \textbf{0.920} & \textbf{4.423} & \textbf{0.905} \\ \midrule
     & GRU w/o sq & & \checkmark & & 3.526 & 0.928 & 4.304 & 0.883 & 4.670 & 0.874 & 6.034 & 0.778 & 6.980 & 0.711\\
     & Transformer & & \checkmark & & 3.211 & 0.934 & 3.985 & 0.909 & 4.733 & 0.881 & 6.287 & 0.790 & 6.526 & 0.710 \\ 
     & *Proposed w/o sq & & \checkmark & & 2.458 &  0.951 & 3.931 & 0.909 & 5.621 & 0.810 & 5.998 & 0.794 & 6.802 & 0.720\\ 
     2018/30th & GRU & & \checkmark & \checkmark & 3.384 & 0.931 & 4.250 & 0.884 & 4.654 & 0.884 & 6.070 & 0.773 & 7.209 & 0.699\\
     -- & ARGO & & \checkmark & \checkmark & 4.034 & 0.920 & --- & --- & --- & --- & --- & --- & --- & --- \\
     2019/29th & Two-stage & & \checkmark & \checkmark & 3.222 & 0.934 & 4.022 & 0.899 & 5.044 & 0.834 & 6.174 & 0.780 & 6.516 & 0.715\\
     & *Proposed\_single & & \checkmark & \checkmark & \textbf{2.390} & \textbf{0.962} & 4.004 & 0.908 & 5.878 & 0.808 & 6.134 & 0.783 & 6.802 & 0.731\\
     & MTEN & \checkmark & & \checkmark & 5.023 & 0.863 & --- & --- & --- & --- & --- & --- & --- & --- \\
     & GRU\_multi & \checkmark & \checkmark & \checkmark & 3.913 & 0.905 & 4.567 & 0.858 & 4.988 & 0.823 & 6.001 & 0.769 & 6.949 & 0.705 \\
     & *Proposed\_multi2 & \checkmark & \checkmark & \checkmark  & 2.800 & 0.955 & 3.557 & \textbf{0.927} & \textbf{4.790} & \textbf{0.868} & 5.694 & 0.814 & 6.483 & 0.789\\
     & *Proposed\_multi5 & \checkmark & \checkmark & \checkmark & 2.897 & 0.952 & \textbf{3.556} & 0.926 & 4.800 & 0.854 & \textbf{5.673} & \textbf{0.815} & \textbf{6.258} & \textbf{0.793} \\ \midrule
     & GRU w/o sq & & \checkmark & & 3.429 & 0.709 & 5.821 & 0.164 & 7.128 & -0.252 & 7.436 & -0.362 & 7.378 & -0.340 \\ 
     & Transformer & & \checkmark & & 3.567 & 0.723  & 5.746 & 0.301 & 7.085 & -0.320 & 7.515 & -0.489 & 7.929 & -0.702\\ 
     & *Proposed w/o sq & & \checkmark & & 3.054 & 0.784 & 4.938 & 0.398 & 6.917 & -0.179 & 7.005 & -0.221 & 7.322 & -0.317\\ 
     2019/30th & GRU & & \checkmark & \checkmark  & 3.303 & 0.722 & 5.655 & 0.297 & 7.360 & -0.357 & 8.085 & -0.514 & 8.622 & -0.729\\
     -- & ARGO & & \checkmark & \checkmark & 3.411 & 0.740 & --- & --- & --- & --- & --- & --- & --- & ---\\
     2020/29th & Two-stage & & \checkmark & \checkmark & 3.569 & 0.731 & 4.585 & 0.401 & 6.812 & -0.166 & 7.383 & -0.333 & 7.890 & -0.586\\
     & *Proposed\_single & & \checkmark & \checkmark & 3.326 & 0.726 & 4.251 & 0.419 & 6.600 & 0.005 & 7.240 & -0.290 & 7.292 & -0.311\\
     & MTEN & \checkmark & & \checkmark & 3.922 & 0.563 & --- & --- & --- & --- & --- & --- & --- & --- \\
     & GRU\_multi & \checkmark & \checkmark & \checkmark & 2.916 & 0.812 & 3.930 & 0.610 & 5.488 & 0.332 & 6.008 & 0.108 & 7.006 & -0.153 \\
     & *Proposed\_multi2 & \checkmark & \checkmark & \checkmark  & 2.936 & 0.830 & 3.724 & 0.649 & 5.400 & 0.312 & 6.135 & 0.072 & 6.972 & -0.197\\
     & *Proposed\_multi5 & \checkmark & \checkmark & \checkmark  & \textbf{2.800} & \textbf{0.858} & \textbf{3.566} & \textbf{0.715} & \textbf{5.004} & \textbf{0.407} & \textbf{5.839} & \textbf{0.282} & \textbf{6.849} & \textbf{-0.121}\\
     \bottomrule
  \end{tabular}
  }
  \\\textbf{*} indicates the variation in the proposed model. Bold indicates the best score in each metric and each term.
    \label{influ_result_japan}
\end{table*}

\section{Experimental results for JP}
The result for JP is presented in Tables~\ref{influ_result_japan}.
This result indicates that the proposed model (particularly our multi-task model) outperformed most baseline methods, confirming the benefits of the model architecture and multi-task learning, same as the US.
The proposed models (\textbf{Proposed\_single}, \textbf{Proposed\_multi2}, and \textbf{Proposed\_multi5}) achieved the best scores with respect to the terms, metrics, and any-ahead forecasts.
These results reveal that the architecture in the proposed model is useful for flu forecasting.
\textbf{Proposed\_single} achieved the best score among the models without multi-task learning in almost all terms except for 2018 to 2019 in JP, in which it exhibited the best score in the near-ahead forecast, whereas it had a lower score in the far-ahead forecast than the GRU-based models.
Same as the US, the high degree of the score improvement in \textbf{Proposed\_multi2} and \textbf{Proposed\_multi5} compared to Proposed\_single demonstrated the usefulness of the multi-task learning, except in 1-week forecast.

In terms of the comparison of models without and with search queries, the experimental results for the flu forecast in JP indicate that the change from GRU w/o sq to GRU resulted in an average improvement of $-0.026$ points in the $\rm{RMSE}$, and of $-0.030$ points in the $\rm{R^{2}}$.
However, the change from Proposed w/o sq to  Propose\_single resulted in an average improvement of $0.187$ points in the $\rm{RMSE}$, and of $0.012$ points in the $\rm{R^{2}}$.
Same as the US, this suggests that the search query data resulted in the GRU-based models exhibits low or worse improvement scores by adding them, however the proposed model, with a well-crafted architecture for the search query data input, achieved a significantly improved score.

\section{Examples of search queries based on each selection method}
Examples of search queries based on each selection method are presented in Table~\ref{example_q}.
Compared to the translation-based method, the WT-based method exhibited many similar points in the selection results, although several aspects differed.
For example, in Japanese, the abbreviation representation ``\begin{CJK}{UTF8}{ipxm}インフル\end{CJK} (I-N-FU-LU)'' is selected 
as the corresponding word for ``flu,'' and not ``\begin{CJK}{UTF8}{ipxm}インフルエンザ\end{CJK} (I-N-FU-LU-E-N-ZA).''
In French, ``infection'' is selected as the corresponding word for ``symptoms.''

\begin{CJK}{UTF8}{ipxm}
\begin{table}[!tb]
  \centering
    \caption{Examples of selected search queries by translation-based and WT-based methods.}
    \small
    \begin{tabular}{c|c|cc}\toprule
    English query & & Translation-based & WT-based\\ \midrule
    \multirow{2}{*}{fever\textvisiblespace and\textvisiblespace flu}& ja &発熱\textvisiblespace と\textvisiblespace インフルエンザ & 熱\textvisiblespace インフル \\
    & fr & fi\'evre\textvisiblespace et\textvisiblespace grippe & fi\'evre\textvisiblespace grippe\\ \midrule
    \multirow{2}{*}{the\textvisiblespace flu}& ja &インフルエンザ& インフル\\
    & fr & la\textvisiblespace grippe & grippe\\ \midrule
    \multirow{2}{*}{symptoms\textvisiblespace of\textvisiblespace flu}& ja &インフルエンザ\textvisiblespaceの\textvisiblespace症状& 徴候\textvisiblespaceインフル\\
    & fr & sympt\^omes\textvisiblespace de \textvisiblespace la \textvisiblespace grippe& infection\textvisiblespace grippe \\ \bottomrule
\end{tabular}
\\``\textvisiblespace'' indicates space for search.
\label{example_q}
\end{table}
\end{CJK}

\section{Effect of the country embedding}
To examine the country embedding effectiveness, we validated the degree of improvement of the two proposed models without and with country embedding (Proposesd\_multi5 w/o CE and Proposesd\_multi5).
The experimental results for the flu forecast in US and JP are presented in Table~\ref{ce_result}.
Proposesd\_multi5 achieved relatively better scores than those without country embedding in two countries.
This suggests that the country embedding, which is the initial latent representation of two GRUs regarding the time series of the search queries and deseasonalized component for each country, exhibits improvement scores.

\begin{table*}[!tb]
    \centering
    \caption{Comparison of forecasting performances of Proposesd\_multi5 and Proposesd\_multi5 without country embedding (CE) in US and JP from 2017/30th week to 2018/29th week.}
    \small
    \centering
    \begin{tabular}{c|l|rrrrrrrrrr} \toprule
    \multicolumn{1}{c|}{\multirow{2}{*}{Country}} & \multicolumn{1}{c|}{\multirow{2}{*}{Method}} & \multicolumn{2}{c}{1-week} & \multicolumn{2}{c}{2-week} & \multicolumn{2}{c}{3-week} & \multicolumn{2}{c}{4-week} & \multicolumn{2}{c}{5-week}\\ \cmidrule(lr){3-4} \cmidrule(lr){5-6} \cmidrule(lr){7-8} \cmidrule(lr){9-10} \cmidrule(lr){11-12}
    & & $\rm{RMSE}$ & $\rm{R^{2}}$ & $\rm{RMSE}$ & $\rm{R^{2}}$ & $\rm{RMSE}$ & $\rm{R^{2}}$ & $\rm{RMSE}$ & $\rm{R^{2}}$ & $\rm{RMSE}$ & $\rm{R^{2}}$ \\ \midrule
    \multirow{2}{*}{US} & Proposesd\_multi5 w/o CE & 0.299 & 0.952 & 0.535 & 0.935 & 0.788 & 0.832 & 0.902 & 0.800 & 0.994 & 0.743 \\
    & Proposesd\_multi5  & \textbf{0.237} & \textbf{0.986} & \textbf{0.498} & \textbf{0.941} & \textbf{0.692} & \textbf{0.837} & \textbf{0.805} & \textbf{0.832} & \textbf{0.942} & \textbf{0.770} \\ \midrule
    \multirow{2}{*}{JP} & Proposesd\_multi5 w/o CE & \textbf{2.400} & \textbf{0.970} & 3.367 & 0.939 & 3.701 & 0.933 & 4.449 & 0.909 & 5.102 & 0.875\\
    & Proposesd\_multi5 & 2.429 & \textbf{0.970} & \textbf{2.878} & \textbf{0.951} & \textbf{3.411} & \textbf{0.941} & \textbf{4.057} & \textbf{0.920} & \textbf{4.423} & \textbf{0.905}\\ 
    \bottomrule
    \end{tabular}
    \label{ce_result}
\end{table*}

\section{Effect of the attention in the proposed model}
\begin{figure}[b]
        \includegraphics[width=\textwidth]{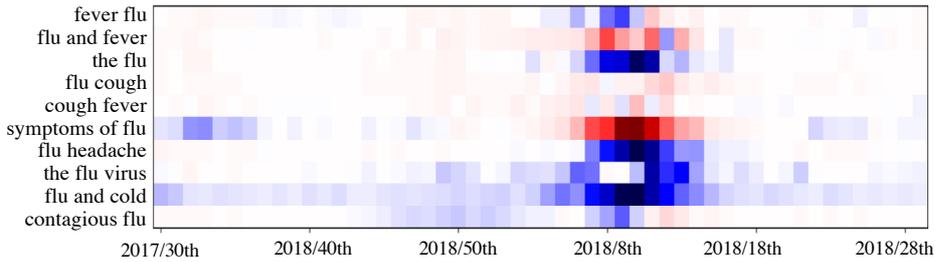}
        \caption{Visualization of attention weight of each search query for forecast in US from 2017/30th week to 2018/29th week. Red indicates a large attention weight, whereas blue indicates a small attention weight.}
        \label{attention}
\end{figure}

The attention mechanism in the proposed model not only successfully combines the search query data, but can also provide a broad understanding of which queries affect the forecast and in what manner.
Fig.~\ref{attention} presents the visualization of the attention weight of each search query for the forecast in the US from 2017/30th week to 2018/29th week.
A large change occurred in the attention weight around 2018/8th week, at which time the flu became an epidemic, whereas the attention weight in each query was almost constant except during this period.
This means that the information from search queries is useful in forecasting the flu during an epidemic period.
In particular, ``flu and fever'' and ``symptoms of flu'' are useful in the flu forecasting model because they have a large attention weight.
It is interesting that the weight of ``flu and fever'' was large despite that of ``fever flu,'' with the same meaning, being small.
Using attention to visualize useful search queries offers the potential to aid in determining the resources of the input for creating the forecasting model in situations where a useful search query has not been determined.

\end{document}